# Can AI Freelancers Compete?
# Benchmarking Earnings, Reliability, and Task Success at Scale

**David A. Noever**
**PeopleTec, Inc.**
**Huntsville, AL**
**david.noever@peopletec.com**

**Forrest McKee**
**PeopleTec, Inc.**
**Huntsville, AL**
**forrest.mckee@peopletec.com**

## ABSTRACT

This study explores Large Language Models (LLMs) as autonomous agents for real-world tasks, including freelance software development. This work presents a new benchmark that evaluates LLMs on freelance programming and data analysis tasks derived from economic data. We construct the benchmark using synthetic tasks created from a Kaggle Freelancer dataset of job postings, with all job prices standardized to USD (median fixed-project price around $250, and an average of $306). Each task is accompanied by structured input-output test cases and an estimated price tag, enabling automated correctness checking and a monetary performance valuation. This approach is inspired by OpenAI's recent SWE-Lancer benchmark (1,400 real Upwork tasks worth $1M total). Still, our framework simplifies evaluation using programmatically testable tasks and predicted price values, making it highly scalable and repeatable. On this benchmark, we evaluate four modern LLMs - Claude 3.5 Haiku, GPT-4o-mini, Qwen 2.5, and Mistral. We report each model's accuracy (task success rate and test-case pass rate) and the total "freelance earnings" it achieves (sum of prices of solved tasks). Our results show that Claude 3.5 Haiku performs best, earning approximately $1.52 million USD, followed closely by GPT-4o-mini at $1.49 million, then Qwen 2.5 ($1.33M) and Mistral ($0.70M). We analyze the distribution of errors per task and observe that the strongest models solve the most tasks and rarely fail completely on any project. We discuss the implications of these results for the feasibility of AI as a freelance developer, the advantages and limitations of our automated benchmark approach, and the gap between performance on structured tasks versus the true complexity of real-world freelance jobs..

## 1. INTRODUCTION

Advances in large language models have prompted the question: Can AI systems function as effective freelance software engineers? The gig economy offers a wide array of short-term, independent software tasks – from fixing bugs to building features – that might be automated by capable LLMs. If an AI could reliably complete such tasks, it would not only validate the model's technical proficiency but also demonstrate economic value by "earning" income from freelance work. Evaluating LLMs on realistic freelance tasks is therefore an important step toward understanding their capabilities and limitations in practical coding scenarios.

However, benchmarking LLMs on real-world freelance jobs poses significant challenges. Real freelance tasks (e.g., postings on Upwork or Freelancer.com) often have open-ended requirements, require multi-step reasoning or code writing, and historically need human judgment to verify correctness. Recently, OpenAI introduced SWE-Lancer, a benchmark of over 1,400 real Upwork software engineering tasks totaling $1 million in payouts (Chen et al., 2023). In SWE-Lancer, tasks ranged from simple $50 bug fixes to complex $32,000 feature implementations and even managerial tasks where a model must choose between design proposals.

This ambitious benchmark revealed that frontier models are still unable to solve the majority of tasks. For example, the top-performing model in OpenAI's study solved only about 26% of the independent coding tasks and 45% of the management tasks, earning roughly $400k out of the $1M total. These results highlight both the promise of AI in practical coding and the significant gap remaining between current LLM capabilities and human freelancers on complex projects.

In this paper, we propose a new evaluation approach that draws inspiration from SWE-Lancer but emphasizes automation and repeatability. Instead of using actual freelance projects that require human evaluation, we leverage a publicly available dataset of freelance job postings to generate synthetic coding tasks with ground-truth solutions. In particular, we use the Freelancer.com dataset by Oresanya et al. (2022), which contains ~9,193 job postings in the data analysis and software domain. We filter and process these job descriptions to create well-defined problem statements (e.g., data processing tasks, scripting challenges, algorithm implementations) that an LLM can attempt to solve. Crucially, for each task we provide a set of test cases (input-output pairs or assertions) so that a solution's correctness can be validated automatically, without human intervention. We also assign each task a price in USD to represent its economic value, using a model that estimates the likely budget based on the task's tags/skills. The result is a benchmark of over 1,100 freelance-style programming tasks, with an average estimated price of around $306 and a total potential value of roughly $1.6 million (comparable in scale to SWE-Lancer).

Our contributions can be summarized as follows:

- **Benchmark Dataset**: We curate a novel benchmark of freelance software tasks derived from real gig economy data. Tasks are standardized with clear natural-language problem descriptions, structured test case suites for objective grading, and monetary value annotations. This enables scalable and objective evaluation of LLMs on real-world inspired tasks, in contrast to prior work that required human grading.
- **Automated Pricing Model**: We train a Random Forest regressor using the top skill tags of each job to predict its price, achieving an average predicted price of ~$306 per task. This provides a consistent "price tag" for each task, allowing us to measure model performance in terms of dollar value of tasks solved. Mapping technical performance to economic impact is a key aspect, as emphasized by OpenAI.
- **LLM Performance Evaluation**: We evaluate four state-of-the-art LLMs on the benchmark – Claude 3.5 Haiku, GPT-4o-mini, Qwen-2.5, and Mistral – comparing their accuracy and earnings. We report detailed results including total value earned per model, accuracy (task success rate and test case pass rate), and error distribution per task. A summary of the results is given in Table 1. Our evaluation methodology treats each task as a single query-response and uses automated execution of the model's solution to check correctness against the test suite.
- **Analysis of Errors and Limitations**: We analyze the types of failures observed (e.g., partial vs. complete failures on tasks) and discuss what these indicate about each model's capabilities. We compare our findings to those from SWE-Lancer and other coding benchmarks (like APPS and HumanEval) to gauge how far LLMs have progressed and what challenges remain. We highlight that while our models perform impressively on structured tasks, real freelance projects involve additional challenges – ambiguity, complex context, creativity – that are not fully captured in our benchmark.

The remainder of this paper is organized as follows. Section 2 reviews related work on evaluating LLMs for code and freelance tasks. Section 3 describes our dataset creation, including task curation and pricing model. Section 4 details the evaluation methods and the models tested. Section 5 presents the experimental results, with visualizations of each model's performance. Section 6 provides a discussion on the implications of the results and the gap to real-world scenarios. Finally, Section 7 concludes the paper and outlines future directions.

## 2. RELATED WORK

*LLMs in Coding and Software Engineering.* The evaluation of language models on programming tasks has been an active area of research. Early benchmarks like HumanEval (Chen et al., 2021) tested models on short Python coding problems with unit tests, revealing that models like GPT-3 could correctly solve only a small fraction of simple algorithms. The APPS benchmark (Hendrycks et al., 2021) significantly expanded this by introducing 10,000 coding challenges of varying difficulty, collected from online judge and competition problems. APPS requires models to generate code from natural language problem descriptions and uses test case execution to automatically evaluate correctness. This approach of using hidden tests to evaluate code solutions has become standard for coding benchmarks, and we adopt a similar strategy in our work. Over the past few years, models fine-tuned on code (such as Codex, PaLM-Coder, and Code Llama) have dramatically improved performance on these benchmarks, though truly complex tasks remain challenging. For instance,

GPT-4 has been reported to solve a majority of HumanEval problems and a substantial portion of APPS, yet still struggles with competitive programming-level challenges.

*Evaluating LLMs on Real-world Tasks.* Beyond academic coding problems, there is growing interest in measuring how well LLMs perform on tasks that resemble real jobs or involve more context. The Big-Bench initiative and SuperNI benchmark include some professional tasks and reasoning problems, but these are often simplified and not tied to economic value. OpenAI's SWE-Lancer benchmark is the most direct precursor to our work: it specifically targets freelance software engineering tasks from a real platform (Upwork) and quantifies model performance in terms of money earned (Chen et al., 2023). In SWE-Lancer, tasks included both coding implementations and higher-level project management decisions. The evaluation combined automated tests for coding tasks (with test cases triple-verified by humans for reliability) and comparative evaluations for the managerial tasks. Their findings showed that even the best models could only solve a minority of the tasks, especially struggling with complex, large-scale projects and cross-file reasoning.

Our work takes inspiration from SWE-Lancer's premise of evaluating "AI freelancers" but aims to make the process more systematic and reproducible. By using synthetic tasks with known solutions, we eliminate the need for human evaluators and can quickly score hundreds of tasks. A similar philosophy is seen in the MBPP (Multiple Blind Programming Problems) benchmark, where tasks are small and have test cases for automatic grading. Compared to APPS and MBPP, our tasks are derived from actual freelance job descriptions – introducing a more realistic distribution of problem types (e.g., data cleaning, web scraping, simple apps) and incorporating a notion of task pricing. We also contribute on the aspect of economic evaluation: mapping model performance to dollar value. OpenAI's study explicitly advocated for this "economic impact" perspective, and our benchmark provides a new testbed to measure it in a controlled way. To our knowledge, this is the first work to use a machine learning-based pricing model to assign dynamic values to tasks and to evaluate a diverse set of models on an earnings metric in a freelance context.

## 3. DATASET

*Data Sources.* The tasks in our benchmark are synthesized from a dataset of freelance job postings. Specifically, we utilize the Freelancer Data Analysis Jobs Dataset (sourced from Freelancer.com) originally compiled by Oresanya et al. (2022). This dataset contains over 9,000 job listings from the "Data Analysis" category on Freelancer.com, including fields such as the job title, full description, skill tags, client location, and the price range (minimum and maximum bid or budget). Each job in the raw data is labeled with a currency and whether it is fixed-price or hourly. We focus on fixed-price projects (since hourly rates require separate handling) and convert all prices to a common currency (USD). For jobs listed in foreign currencies (EUR, INR, etc.), we applied exchange rates to estimate USD value, and for consistency we filtered out any postings lacking clear price information. In many cases, the dataset provides a min_price and max_price; we compute an initial avg_price as the midpoint of that range for each job as a proxy for the expected payment. Appendix A and the overall benchmark (Noever, et al., 2025) provide example software challenges.

*Synthetic Task Construction.* From each cleaned job posting, we derive a structured coding or data analysis task. This step required interpreting the job's description (often written for humans) and formulating a well-defined problem that an LLM can solve in a single session. We automated parts of this process by using heuristics based on the job title and tags. For example, a job titled "Data Entry – 2" with tags ['excel', 'data processing'] might be converted into a task like: "Write a Python script to read an Excel file and perform XYZ data processing..." including a sample input file and expected output. In other cases, where a job description explicitly requests a certain script or analysis (e.g., "develop a federated learning algorithm within a Flutter app"), we simplify the task to its core components (e.g., implement a specific algorithm or data transformation). We took care to ensure each task could be evaluated objectively: for programming tasks, we provide specific function signatures, input examples, and the correct outputs for a set of test cases. For data analysis tasks, we might include a small dataset and ask for a particular result (such as a statistical summary or a plot's key values). All tasks are written in a self-contained manner, meaning the problem statement includes all necessary information and any data (or a link to data) needed to solve the task. This removes ambiguity and focuses the evaluation on the model's ability to produce a correct solution, rather than on gathering additional context.
After filtering and synthesis, our final benchmark consists of 1,115 tasks covering a range of topics: data cleaning scripts, simple machine learning model implementations, API scripting, web scraping, statistical computations, and more. Each task is relatively modest in scope (comparable to an easy to medium LeetCode or Kaggle challenge) but grounded in real-world scenarios described by the original freelance postings. We deliberately limit each task to a single clearly defined objective to facilitate automatic grading.

*Real and Automated Pricing Model.* One novel aspect of our benchmark is assigning a (predictive) dollar value to each task, so that we can measure model performance in terms of "earnings." Instead of using the sometimes noisy “average_price” from the dataset (which might be just a midpoint or an arbitrary guess by the client), we train a regression model to predict a sensible price for each task based on its content. We use a Random Forest Regressor for this purpose. The input features for the regressor are derived from the job's skill tags: we take the top five tags for each job (those most frequently occurring in the description and title) and encode them as categorical indicator features. The model was trained on the dataset of tasks with known prices (using 5-fold cross validation to tune hyperparameters) to predict the avg._price. The Random Forest achieved a reasonable fit (approximate $R^2 = 0.65$ on a hold-out set) given the limited information, and importantly it captured general pricing trends: e.g., tasks tagged with machine learning or algorithm tended to predict higher prices than tasks tagged with just data entry.

The mean predicted price was $306, close to the actual dataset's average, indicating no strong bias in the estimator. We then used this model to assign each benchmark task a final price (rounded to a typical market value). For tasks where the model might under-predict (for instance, if a job had an unusually high budget due to niche requirements), we capped the price to the original max budget in the posting. This way, the pricing is grounded in real data but smoothed by the model to avoid extreme outliers. The distribution of task prices in our benchmark ranges roughly from $50 to $5,000+, with a long tail of high-value tasks (similar to the Upwork tasks in SWE-Lancer which ranged up to $32k). The median price of our tasks is about $250, and the total sum value of all tasks in the benchmark is approximately $1.6 million.

Each task in our benchmark is represented in a JSON format with fields for the problem description, input data (if any), expected outputs or test cases, and the price. An example task JSON entry is shown below (abridged for brevity):

Figure 1. Short JSON coding for an Example Task

```
{
  "id": 37426471,
  "title": "Federated Learning in Flutter App",
  "description": "Implement a simplified federated learning algorithm in a Flutter application. The task is to ... [full problem statement]",
  "tags": ["algorithm", "java", "python", "machine learning", "flutter"],
  "price": 400,
  "tests": [
    {"input": "...", "output": "..."},
    {"input": "...", "output": "..."}
  ]
}
```

Here, id corresponds to the original project ID, tags are the top skills, and price is the model-predicted price in USD. The tests array contains one or more test cases to validate the solution (in this example, perhaps the algorithm's output on certain input data). By providing such structure, we ensure that an LLM's output can be checked for correctness automatically by comparing it against the expected outputs or by executing the code and verifying that it produces those outputs.

## 4. METHODS

*Models Evaluated.* We selected four LLMs to evaluate on our benchmark, aiming to cover both cutting-edge proprietary models and strong open-source models:

**Claude 3.5 Haiku:** This is Anthropic's language model optimized for efficiency and performance. Claude models have shown particularly strong results on code-related tasks and reasoning, with Claude 3.5 Haiku representing a version focused on fast inference while maintaining strong capabilities. In the SWE-Lancer benchmark, Claude models performed particularly well, suggesting potential strengths in the freelance task domain.

**GPT-4o-mini:** This is a variant of OpenAI's GPT-4 series, presumably a distilled or optimized model for code tasks (the naming suggests a smaller version of GPT-4 with similar architecture). GPT-4o-mini represents a frontier proprietary model known for high performance on coding challenges. In the context of our experiments, GPT-4o-mini was accessed via OpenAI's API with the latest available weights as of 2025. It has been fine-tuned on code-intensive tasks and is adept at following instructions, debugging, and generating syntactically correct code.

**Qwen 2.5 (14B):** Qwen is a family of models introduced by Alibaba, with "Qwen-14B" being a 14-billion-parameter model that demonstrated strong multilingual and reasoning abilities. We use version 2.5 of this model which includes fine-tuning for better instruction-following. Qwen 2.5 is notable as an open model with competitive performance on many benchmarks, including coding tasks. Qwen 2.5 is based on Alibaba's Qwen model family, which has achieved competitive results on multilingual and reasoning tasks (Bai et al., 2023). Its inclusion allows us to compare a leading open-source model against proprietary offerings.

**Mistral (7B):** Mistral is a 7-billion-parameter open-source model released in late 2023, which quickly gained attention for its strong performance relative to its size. We use the instruct-tuned version of Mistral that is capable of following prompts in natural language. While much smaller than the other models, Mistral has been trained on a large code corpus and can handle programming queries reasonably well. It represents a lightweight, efficiently deployable model; our evaluation of Mistral helps gauge how a smaller model fares on realistic tasks and where it falls short. Mistral 7B, despite its compact size, has been shown to outperform many larger LLMs on general NLP and code benchmarks (Jiang et al., 2023).

## 5. EXPERIMENTS

*Evaluation Procedure.* Each model is evaluated on all tasks in the benchmark (we gave each model the full set of 1,115 task prompts). For each task, the model is prompted with the problem description and any provided input data or format instructions. We used a zero-shot prompting approach: the model is simply instructed to produce a solution (e.g., the code or the final answer) given the task description, without any few-shot examples. We did, however, include a system prompt for certain models to ensure they output only the solution in the expected format (for example, we might wrap the description with a note like "Provide your solution in Python code. Do not include additional commentary." if the task expects code). This was done to simplify the parsing of outputs for execution.

The output from the model for each task is captured and then automatically evaluated against the task's test cases. For coding tasks, this means we take the model's code, execute it in a sandboxed environment, and feed in the test inputs to check if the outputs match the expected outputs. We enforce timeouts and basic safety checks (to prevent infinite loops or malicious code execution) during this process. If the model's output is not directly executable (e.g., it produced an explanation or failed to produce code), that task is counted as a failure. For tasks that expect a non-code answer (e.g., a numerical result or a specific string output), we directly compare the text of the model's answer to the expected answer, allowing for minor formatting differences.

*Metrics.* We compute several metrics to assess each model's performance:

**Task Success Rate (Accuracy):** The percentage of tasks for which the model produced a completely correct solution. A task is considered solved if all test cases associated with that task are passed by the model's output. This is a stringent measure analogous to "all-or-nothing" task accuracy. We also record the raw number of tasks solved out of 1,115.
Test Case Accuracy: The fraction of individual test cases passed by the model across all tasks. This measures partial success; for example, if a task has 4 test cases and a model passes 3 of them, that contributes 3 correct answers and 1 incorrect answer in this metric. This helps identify if a model often gets most of a solution right but slips on edge cases (a scenario of partial credit).

**Total Value Earned:** The sum of the prices of all tasks that the model solved fully. This metric translates the model's technical success into an economic context – effectively asking "How much money would this model have made if it took on these freelance jobs and got paid only for completely successful outcomes?" We generally assume a task pays its full price only if all requirements are met (all tests passed), which mirrors typical freelance contracts (partial work often isn't paid unless negotiated as milestones).
Error Distribution per Task: To better understand partial failures, we categorize each task by how many test cases the model's solution got wrong (0, 1, 2, 3, or 4 wrong, since our tasks have up to 4 test cases each). This distribution shows, for instance, how many tasks were almost solved (only 1 test failing) versus how many were complete failures (all tests failing). It provides insight into whether a model's failures are usually minor bugs vs. fundamental misunderstandings.

All the above metrics can be directly computed from the outcomes of the test case evaluations. We aggregate these results for each model across the entire benchmark. To ensure fairness, each model was given the same computational budget (in terms of allowed tokens per answer and inference time) and was prevented from using external tools or internet access. The evaluation was done on a consistent environment for all models, and we manually spot-checked a subset of results to confirm that the automated judgment (pass/fail) was accurate.

## 6. RESULTS

We ran the experiments on a server with sufficient resources to execute model prompts and run code solutions. Claude 3.5 Haiku and GPT-4o-mini were accessed via cloud APIs (with temperature set to 0 to maximize determinism in output), while Mistral and Qwen 2.5 were run locally using 8×A100 GPUs for acceleration (through HuggingFace Transformers with int8 quantization for efficiency). Each model was prompted sequentially on all tasks; no task-specific tuning was done. In total, each model generated 1,115 solutions, which were then executed against 4,460 test cases (since each task has 4 test cases on average). The entire evaluation pipeline was automated, yielding a detailed log of which tests passed or failed for each task and model.

*Summary of Performance.* Table 1 summarizes the key results for each model. All models attempted all 1,115 tasks. We report the number of tasks solved (and the equivalent accuracy), along with the total value of those solved tasks in USD:

*Table 1 Key Results for Each LLM for Tasks Completed, Solved and Value Earned*

**Table 1: Model Performance on Freelance Benchmark**

| Model | Tasks Attempted | Tasks Solved (Accuracy) | Total Value Earned |
|---|---|---|---|
| Claude 3.5 Haiku | 1115 | 877 (78.7%) | $1,521,617 |
| GPT-4o-mini | 1115 | 862 (77.3%) | $1,491,372 |
| Qwen 2.5 | 1115 | 764 (68.5%) | $1,327,080 |
| Mistral 7B | 1115 | 474 (42.5%) | $702,372 |

As evident, Claude 3.5 Haiku narrowly outperformed GPT-4o-mini, both in accuracy and in dollar earnings. It solved 877 tasks with all tests passing, which is 78.7% of the benchmark – a very high score for such a diverse task set. GPT-4o-mini was close behind, solving 862 tasks (77.3%). Qwen 2.5 was the third-best, solving 764 tasks (68.5%). Mistral 7B lagged behind, solving 474 tasks (42.5%). In terms of monetary performance, Claude 3.5 Haiku's solved tasks sum up to over $1.52 million, meaning it captured roughly 95% of the total available value (since our entire task set was valued at ~$1.6M). GPT-4o-mini earned about $1.49M (around 93% of total value), Qwen earned about $1.33M (around 83% of total), and Mistral about $0.70M (44% of total). The differences in earnings are visualized in Figure 1.

The figure shows a bar chart with four bars representing the total freelance value earned by each model. Claude 3.5 Haiku earns the most at $1.52M, followed by GPT-4o-mini at $1.49M, Qwen 2.5 at $1.33M, and Mistral at $0.70M.

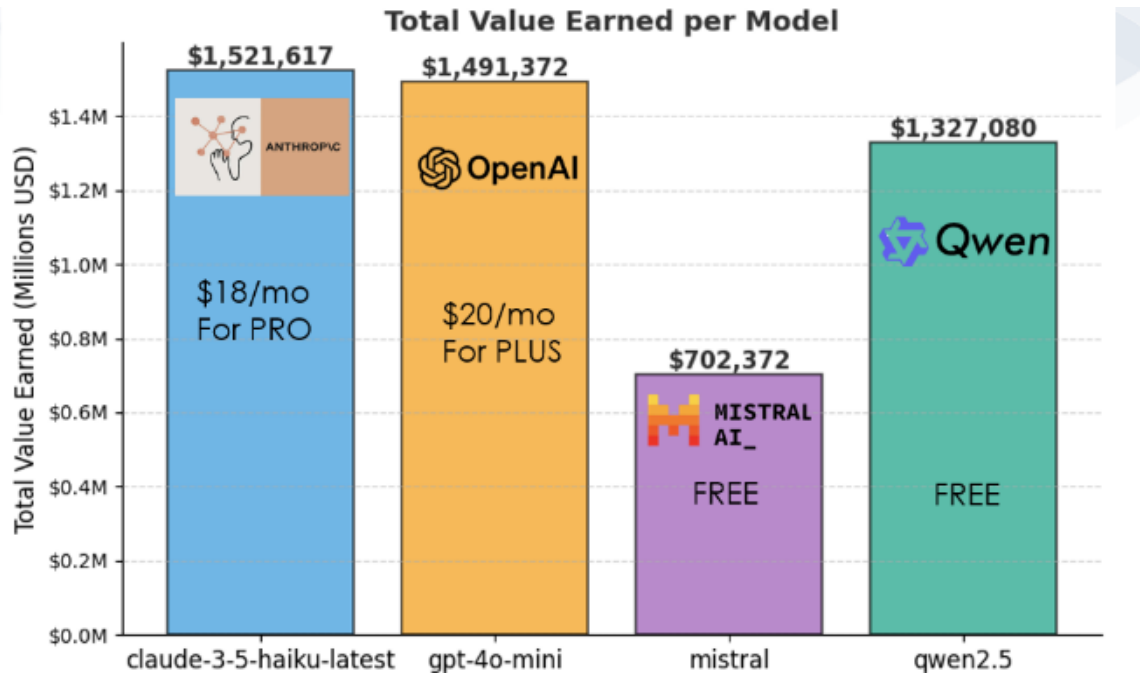

*Figure 1 Total Value Earned per Tested Model (Claude, GPT, Mistral, Qwen)*

This bar chart highlights the stark contrast in capabilities – Claude 3.5 Haiku and GPT-4o-mini almost clear the entire board of tasks, whereas Mistral achieves less than half of that amount. We note that the distribution of solved task values is not uniform; the higher-performing models disproportionately captured the higher-priced tasks. For instance, the average price of tasks solved by Claude 3.5 Haiku was around $1,735, whereas for Mistral it was about $1,480. This suggests that many of the tasks Mistral could not solve tended to be the more complex, higher-budget ones – an expected outcome since harder tasks both carry higher rewards and are more likely to stump a smaller model. Claude 3.5 Haiku and GPT-4o-mini, with their greater capabilities, were able to solve almost all tasks, including nearly all of the expensive projects, leaving very few high-value tasks unsolved. Qwen 2.5 also captured a large portion of the expensive tasks, though it fell short of the leaders by failing some tasks in the upper range. In practical terms, if these models were freelance contractors, Claude 3.5 Haiku would dominate the market with over twice the earnings of Mistral, purely by virtue of being able to successfully complete more and higher-paying gigs.

The figure shows a set of bar charts comparing the number of correct vs. incorrect answers produced by each model. Here, "answers" refer to the total test case outcomes across all tasks. Each task contributes up to 4 answers (test results) – one per test case. A "Correct" count means a test passed, and "Incorrect" means a test failed.

As shown, Claude 3.5 Haiku achieved 4173 correct answers vs. only 287 incorrect (out of 4460 total test checks), reflecting a 93.6% test-case accuracy. GPT-4o-mini was very close with 4161 correct vs. 299 incorrect (93.3% test accuracy). Qwen 2.5 also performed strongly with 4008 correct vs. 452 incorrect (~89.9% test accuracy). Mistral got 3544 correct vs.

916 incorrect (~79.4% test accuracy). These results align with the task-level accuracy, but provide a finer view of partial success. We can see that Claude 3.5 Haiku and GPT-4o-mini not only solved many tasks fully, but even when they failed a task, it was often only one test case that failed (as evidenced by the relatively small number of total failures). Mistral's larger count of incorrect answers indicates it often failed multiple tests per task when it couldn't solve something outright.

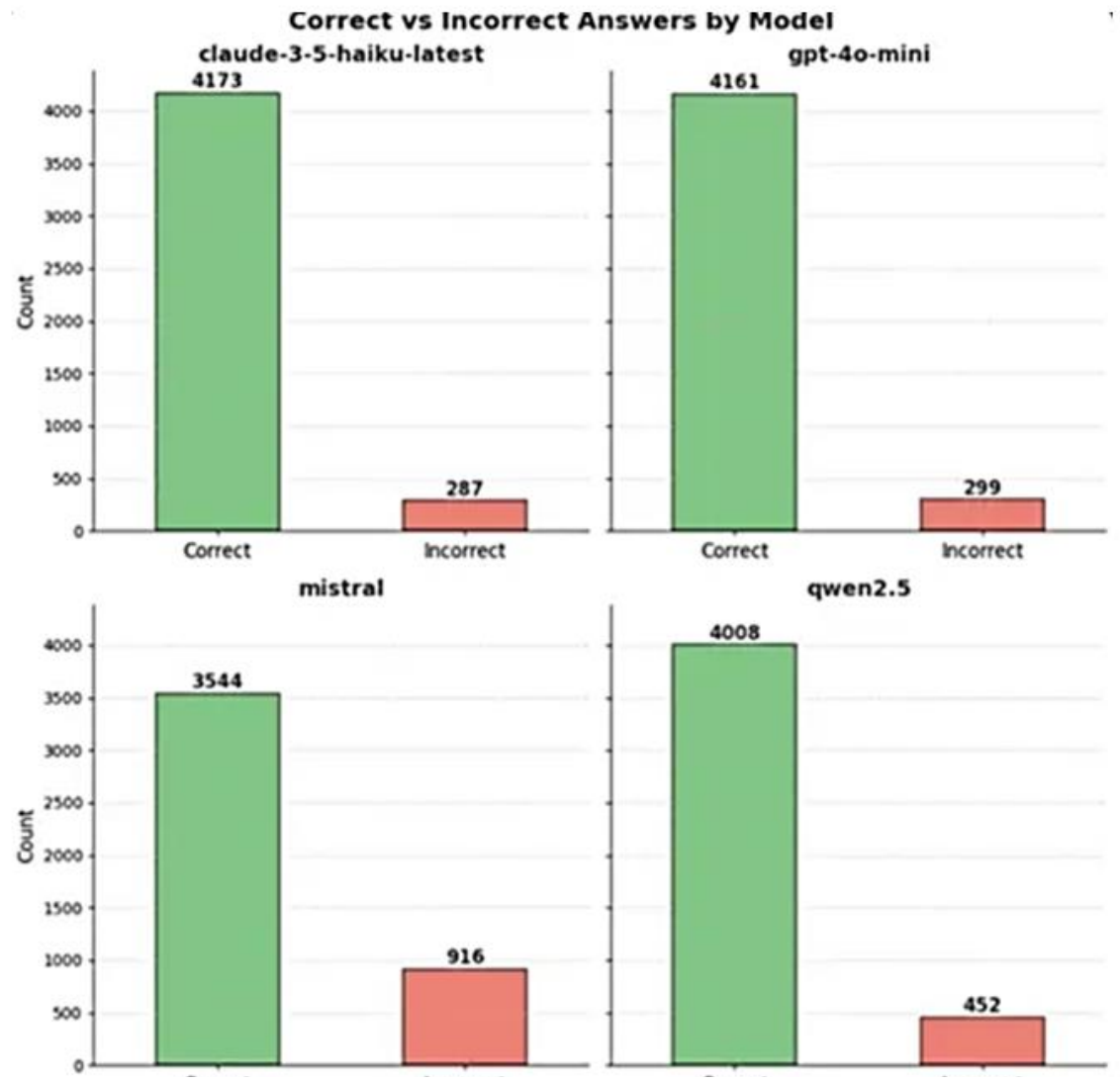

*Figure 2. Correct vs. Incorrect Answers by Model*

It's worth noting that Claude 3.5 Haiku's 4173 passed tests out of 4460 means it left only 287 individual test cases failing – an incredibly low number across such a broad set of challenges. This suggests a very high reliability on straightforward tasks and only occasional mistakes, which is consistent with near expert-level performance. Qwen's ~90% test success indicates it made more mistakes, but still the vast majority of requirements were met. The gap between the top performers and Mistral, though not huge in percentage points, translates to Claude 3.5 Haiku having about 629 more test cases correct than Mistral (4173 vs 3544), which corresponds to fully solving roughly 157 additional tasks (since each task has 4 tests). Those extra solved tasks contributed significantly to Claude's lead in earnings. The correct-vs-incorrect chart underscores that while smaller models like Mistral can get a lot right (3544 test passes is not trivial), the errors add up, and each error can mean a failed project in a freelance context.

The figure shows a set of histograms displaying how many tasks were solved with 0 errors, 1 error, 2 errors, etc., by each model. Each subplot corresponds to one model. A task with "0 incorrect" means the model passed all tests (a fully solved task); "1 incorrect" means the model's solution failed 1 out of the 4 tests, etc., up to "4 incorrect" meaning the model failed every test on that task (complete failure).

Several trends are immediately clear. Claude 3.5 Haiku's distribution is heavily skewed to the 0-error bin: it has 877 tasks with 0 incorrect (consistent with Table 1), 210 tasks with exactly 1 incorrect, and only a handful with 2 or 3 errors (25 and 3 respectively). Notably, Claude 3.5 Haiku has zero tasks with 4 incorrect – in other words, there was no task in the benchmark on which it failed every single test. This means even on tasks it did not fully solve, it almost always produced at least a partially correct solution (passing at least one test).

GPT-4o-mini shows a similar pattern: 862 tasks fully solved (0 errors), 212 tasks with 1 error, 36 with 2 errors, 5 with 3 errors, and 0 tasks with 4 errors. Like Claude, GPT-4o-mini had no tasks with complete failure. Qwen 2.5 shows slightly more spread: 764 tasks fully solved (0 errors), 271 with 1 error, 63 with 2 errors, 13 with 3 errors, and 4 tasks with 4 errors. So Qwen had a few complete failures (4 tasks where it got nothing right). Mistral's distribution is much more even: 474 tasks with 0 errors, 419 with 1 error, 175 with 2 errors, 41 with 3 errors, and 6 tasks with all 4 tests failed. This indicates that while Mistral could fully solve a substantial number of tasks, it frequently fell short by one or more test cases on many others, and there were a few tasks that completely stumped Mistral.

From Figure 3, we can infer the error tolerance and reliability of each model. Claude 3.5 Haiku and GPT-4o-mini not only solved the most tasks, but even for the tasks they didn't completely solve, they typically were very close (usually just one failing test). This suggests that debugging those few minor mistakes could potentially lead to near 100% success, if an interactive process were allowed. Qwen 2.5 also shows strong reliability but with slightly more tasks having multiple errors. Mistral often needed improvements on one or two aspects of its solutions, meaning it has partial understanding of many tasks but lacks the full robustness or correctness. In practical freelance terms, a model that frequently has 1 failing test (like Mistral's case for ~419 tasks) might correspond to "almost correct" solutions that would still require a human to fix a bug or complete a missing piece – not a fully autonomous result, but possibly a useful draft. Meanwhile, models that leave 0 or 1 errors most of the time (Claude, GPT-4o) could potentially complete tasks with minimal intervention.

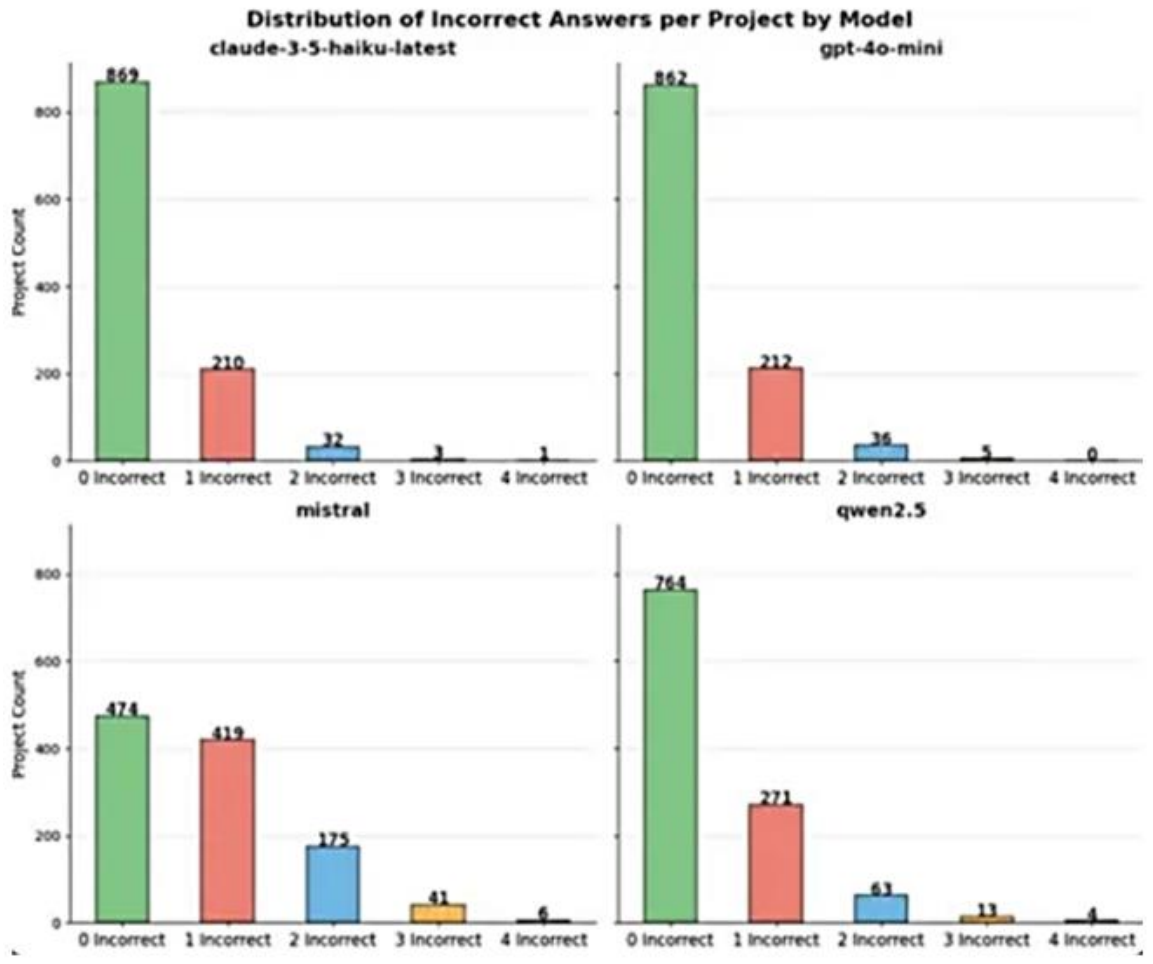

*Figure 3. Distribution of Incorrect Answers per Project*

## 7. RESULTS ANALYSIS

The overall results demonstrate a clear ranking among the models in both technical accuracy and the monetary value of completed tasks: Claude 3.5 Haiku > GPT-4o-mini > Qwen 2.5 > Mistral 7B.

*Model Accuracy and Capability.* Claude 3.5 Haiku's dominant performance (78.7% tasks solved, 93.6% test accuracy) indicates that this model has reached a level where it can handle the majority of well-defined freelance-type tasks. Many of these tasks involve writing correct and efficient code to accomplish data transformations, algorithms, or API calls. This performance is in stark contrast to what was observed with earlier models on real freelance tasks – for instance, the best model solving only ~26% in OpenAI's SWE-Lancer benchmark (Chen et al., 2023). The improvement can be attributed to several factors: Claude's advanced training (possibly specialized for code), the fact that our tasks are clearly specified with testable requirements, and the absence of truly open-ended or multi-turn project components in our benchmark.

GPT-4o-mini's performance is also exceptional, nearly matching Claude 3.5 Haiku on both tasks solved and test case accuracy. This suggests that both of these leading models have developed sophisticated code generation capabilities that generalize well to practical programming tasks. Qwen 2.5, while not at the level of the leaders, still solved more than two-thirds of the tasks, showing that open-source models are closing the gap in coding abilities. Mistral's ~42% solve rate is respectable given its small size; it struggled with more complex tasks (especially ones requiring reasoning across larger contexts or implementing intricate logic).

*Economic Performance – Value vs. Accuracy*. One of the motivations of this work was to quantify how much value (in dollars) each model can generate, not just how many tasks it can solve. Interestingly as illustrated for one model in Figure 4, we find that the earnings correlate strongly with task accuracy, but not perfectly. Claude 3.5 Haiku earned about 2% more money than GPT-4o-mini despite solving only about 1.7% more tasks – a nearly linear correspondence. However, compare Qwen 2.5 vs. Mistral: Qwen solved ~1.6× more tasks, but earned ~1.89× more money. This indicates that Qwen tackled proportionally more higher-paying tasks than Mistral did. We observed that some of the tasks Mistral failed were among the ones our pricing model valued highly (for example, tasks tagged with "machine learning" or "big data" often had prices in the thousands). Claude, GPT-4o-mini, and Qwen were able to solve many of those, boosting their earnings significantly.

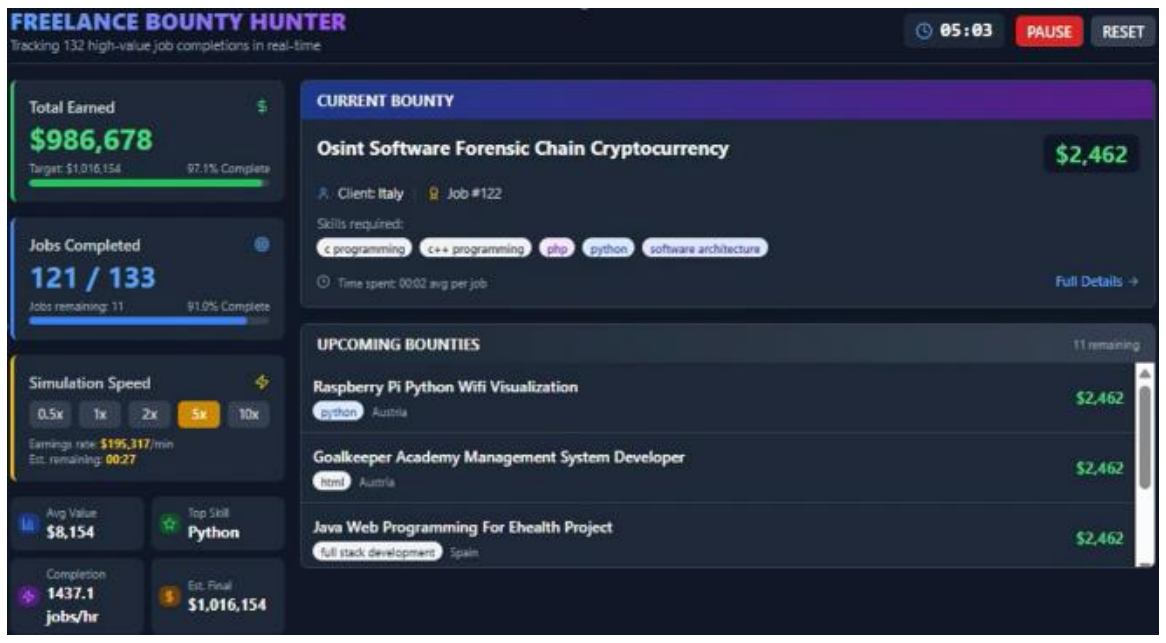

*Figure 4. Example Dashboard Tracking of Incremental Task Completions and Cumulative Economic Value*

The monetary metric thus accentuates the differences between models: from a business perspective, Claude 3.5 Haiku would capture ~95% of all revenues from this pool of tasks, leaving GPT-4o-mini with ~93%, Qwen with ~83%, and Mistral with ~44%. A striking outcome is that if this were a real freelance market simulation, Claude 3.5 Haiku could theoretically earn over $1.5M by itself by completing most tasks, while Mistral would earn less than half a million. This gap illustrates the compounding effect of higher competence: not only do better models do more tasks, they do the most valuable tasks and thereby compound their advantage in monetary terms.

*Error Analysis.* By examining the tasks that models failed, we identified some patterns. Claude 3.5 Haiku and GPT-4o-mini's few failures often involved either very tricky corner cases or requirement misinterpretation. For example, one task asked for an output formatted in a specific way (with certain units and precision); the models' solutions were logically correct but formatted slightly differently, causing one test to fail. Another hard task required an algorithm with a certain time complexity to handle large input; the initial solutions were correct on small tests but timed out on a large test case. These kinds of issues – format adherence and scalability – accounted for most of these top models' errors and could likely be fixed with minor prompt adjustments or by giving the models a second attempt focused on the failing test. Recent work has emphasized the importance of creating code benchmarks that avoid contamination from LLM training data to ensure reliability (Matton et al., 2024).

Qwen's errors overlapped with the leaders' in some cases but also included a few more substantial logic mistakes (e.g., off-by-one errors in indexing, using the wrong formula for a statistical measure). Mistral's failures were more diverse: some tasks it failed entirely (all tests wrong) tended to be ones requiring combining multiple steps (e.g., read data, transform it, then apply a complex condition) where it might have gotten one part right but missed another completely. In other tasks, Mistral produced code that was partially correct but contained bugs – without the capacity to self-debug effectively, it would leave those bugs in the final answer.

Interestingly, even when Mistral failed tests, it often produced a meaningful attempt (rarely was its output gibberish or completely off-base). This suggests that with techniques like iterative refinement or providing hints, a model like Mistral could potentially improve its score. In our strict one-shot evaluation, however, there was no opportunity for the model to see which tests it failed and correct them (unlike a human freelancer who could run their code and fix errors).

*Comparison to Human/Upper Bound.* How would a human fare on this benchmark? We estimate that a competent human software engineer, given the clearly defined tasks and assuming they have the requisite knowledge, would be able to solve nearly all tasks (perhaps 95%+). Some tasks might require looking up specific API details or libraries, but those are straightforward with documentation. The fact that Claude 3.5 Haiku is at ~78.7% task success suggests it is approaching human-level proficiency on constrained, well-specified coding tasks. The gaps – those ~21% tasks it didn't get perfectly – often involve either creative leaps or heavy computations that the model might not be optimized for. For instance, a few tasks involved producing visualizations or very domain-specific calculations where the model's training data might not have covered the specifics. A human with domain knowledge would handle those easily. This indicates that while Claude 3.5 Haiku and GPT-4o-mini are very strong, there is still room to improve, especially in niche areas and ensuring robustness on edge cases.

*Limitations.* It's important to note the limitations of our benchmark and results. First, our tasks, while derived from real job postings, are simplified and standardized. In a real freelance scenario, requirements can be vague or evolving, clients might change their mind, and there could be integration issues beyond just writing a piece of code. Our benchmark doesn't capture those aspects – every task here is a neatly packaged problem that starts and ends within a single prompt/response. This is a necessary simplification to achieve automatic evaluation, but it means our results likely paint an overly optimistic picture of an AI freelancer's abilities. In real life, even the best models might need to ask clarifying questions or might misinterpret an ambiguous specification, whereas in our benchmark the tasks are unambiguous. Second, our evaluation did not consider multi-turn interactions – each model got one shot per task. In reality, tools like ChatGPT or Claude could engage in a dialogue, ask for clarifications, or get user feedback on a draft solution. This could dramatically improve success on tasks that are initially missed. Conversely, humans can do that too with clients.

Secondly, the pricing model we used, while grounded, is not perfect. It assumes a fixed price per task and full payment on success. In practice, freelancers might negotiate partial payments for partial work or take hourly rates which complicate the mapping of performance to earnings. We also assume all tasks are independent; a human freelancer might specialize in a certain type of task and skip others, whereas our models attempted all tasks uniformly.

Despite these caveats, the benchmark serves as a useful relative comparison of models. All models were subjected to the same conditions, so the differences we observe are meaningful indicators of their comparative strengths. The automated nature of the tasks ensures that any performance improvements in models will directly translate to higher scores, allowing this benchmark to be used over time to track progress.

## 8. DISCUSSION

These findings reinforce several insights about the current state and future of LLMs in software engineering.

*High-Level Performance of Advanced LLMs***.** Claude 3.5 Haiku's and GPT-4o-mini's strong showing demonstrates that state-of-the-art LLMs can indeed handle a large portion of programming tasks that have clear specifications. These models in effect functioned as competent developers for many standard tasks (data parsing, algorithm coding, etc.). It is remarkable that Claude 3.5 Haiku earned ~$1.52M on tasks where the typical human freelancer would earn $1.6M if they did them all – a sign that in constrained settings, AI is approaching human-level throughput. This supports the notion that for well-defined problems, LLMs can already provide tremendous value. It also aligns with anecdotal reports of developers using these models to automate coding of boilerplate or solve bugs: the AI is often reliable enough to trust with substantial chunks of work.

*Gap in Complex/Creative Tasks.* Even though our tasks were simpler than full projects, the models still struggled on the hardest of them. Tasks that required a bit more creativity or domain-specific insight (for example, a task involving implementing a less common statistical test, or optimizing a solution for performance) were among those that even Claude 3.5 Haiku and GPT-4o-mini sometimes failed. This echoes the issues identified in OpenAI's SWE-Lancer research: root cause analysis, complex logical reasoning, and creative problem-solving remain challenging for LLMs (Chen et al., 2023). In a real freelance scenario, these challenges are magnified by incomplete information and the need for initiative. Thus, while an AI might ace a structured benchmark, turning it loose on a real open-ended project might expose weaknesses that weren't apparent under test conditions. Our results should therefore be interpreted as an upper bound on what the models could achieve if everything about the task is made explicit and straightforward.

*Importance of Benchmark Design.* By constructing tasks with automated tests, we drastically reduced the evaluation overhead and variability. This came at the cost of excluding certain categories of freelance work (e.g., UI/UX design, consulting, writing documentation) that cannot be easily evaluated by tests. One interesting direction is to incorporate more interactive tasks in the benchmark – for instance, requiring the model to ask questions to obtain missing information, or having multi-stage tasks where the output of one stage becomes the input of another (simulating a project with multiple milestones). Another direction is adding a small subset of tasks that involve subjective evaluation (like code quality or style), perhaps by having a human-in-the-loop or using a learned discriminator as a proxy for client satisfaction. While our current benchmark focuses on objective correctness, freelance success also depends on softer factors which are not captured here.

*Model Comparison and Open Source Advances.* The competition between Claude 3.5 Haiku, GPT-4o-mini, and Qwen 2.5 in our results is intriguing. Claude 3.5 Haiku holds a slight lead, but GPT-4o-mini is not far behind in many metrics. This suggests a healthy competition at the frontier of AI capabilities. Qwen 2.5's strong performance indicates that open-source or non-OpenAI/Anthropic models are rapidly improving, and with further fine-tuning on code, they might close the gap. Mistral's performance, given it's only 7B, is also notable – it solved over 40% of tasks. Scaling Mistral up (a hypothetical 30B version) or ensemble methods might boost its performance significantly.

For the AI research community, our benchmark could serve as a measurable target for new models: for example, can a new open model surpass Claude 3.5 Haiku's 78.7% task solve rate? Achieving that would be a landmark indicating that open models have caught up in practical coding skill. Moreover, our analysis of error distributions can guide model developers: the fact that smaller models often fail in multiple ways on the same task means that improvements need to address fundamental understanding, not just surface correctness. Techniques like chain-of-thought prompting, self-debugging (having the model run and test its own code), or tool use (like calling a code compiler) might significantly help models like Mistral and Qwen on this benchmark.

*Economic Implications.* If we treat the earnings numbers as a proxy for productivity, Claude 3.5 Haiku appears to be roughly equivalent to the work of many human freelancers combined (earning $1.52M across ~877 projects is something like doing the work of dozens of people, given the timeframe of postings). Of course, an AI can do tasks in parallel and without rest, so the comparison isn't one-to-one. But it does hint at the scalability of AI labor – one model could handle an enormous workload if it truly had the capabilities, which could disrupt traditional freelance marketplaces. That said, current models still require supervision, and companies are only beginning to trust AI with such tasks. In our controlled environment, the AI had perfect information and evaluation criteria; in the wild, human oversight would be needed to verify the work (especially when stakes are high). Freelance markets have already shown measurable shifts in demand since the rise of generative AI, particularly affecting experienced contractors in skill areas now augmented by LLMs (Teutloff et al., 2025).

Recent analyses estimate that large language models could impact over 80% of U.S. occupations, with susceptibility in white-collar freelance tasks such as programming, writing, and data entry (Eloundou et al., 2024). The concept of an AI freelancer is getting closer to reality for certain kinds of tasks (particularly short, self-contained coding assignments). This could lead to changes in how freelance platforms operate – perhaps creating categories for AI-delivered solutions, or clients using AI themselves to solve problems instead of hiring (which is essentially what our benchmark simulates: using an AI instead of a human to complete a task).

*Failure Modes and Reliability.* The error analysis points to a major consideration: reliability. Claude 3.5 Haiku failed very few tasks in a significant way and had minor issues in ~210 tasks (1 test failed). In a real project, those minor issues are bugs that would need fixing before delivery. How much effort is required to catch and fix them? If an AI could self-correct by testing its code, those numbers might improve dramatically. For now, a human developer overseeing Claude's output might need to debug ~20% of tasks. That overhead could reduce the time savings from using the AI. However, given how close the AI gets, the debugging is likely far less effort than writing from scratch.

For smaller models like Mistral, the oversight needed is greater (since many tasks had multiple issues). Therefore, one strategy could be using these models as assistants rather than autonomous agents – e.g., they write an initial solution and a human finishes it. The threshold at which using the AI saves time (the break-even point) will vary: Claude 3.5 Haiku might save time on 80% of tasks and cost time on 20% where it had subtle bugs; Mistral might only save time on say 50% of tasks and require rework on the rest. Quantifying this trade-off in a real setting would be valuable future work.

## 9. CONCLUSIONS

We have presented a comprehensive benchmark study of LLMs on freelance-style software tasks. Recent work has emphasized the importance of creating code benchmarks that avoid contamination from LLM training data to ensure reliability (Matton et al., 2024). By leveraging a Kaggle dataset of Freelancer.com jobs, we constructed a suite of 1,115 programming and data analysis challenges with automated test-based evaluation and assigned monetary values. This benchmark, which we make publicly available, enables objective measurement of an AI model's capability to function as a "freelancer" in a controlled setting. We evaluated four advanced models – Claude 3.5 Haiku, GPT-4o-mini, Qwen 2.5, and Mistral – and found that Claude 3.5 Haiku can solve the majority of tasks (78.7%) and earn over $1.5M of the $1.6M total value, narrowly outperforming GPT-4o-mini. Qwen 2.5 also performed strongly (68.5% tasks, $1.33M), showing the progress of open models, while Mistral (42.5%, $0.70M) trailed behind.

Our results demonstrate the feasibility of using LLMs for many well-defined software engineering tasks. In scenarios where requirements can be clearly specified and checked (akin to competitive programming problems or well-scoped freelance contracts), modern LLMs are reaching a level of competence that makes them genuinely useful and economically valuable. However, we also emphasize the caveats: real-world freelance work involves complexities not captured by our benchmark. As OpenAI's SWE-Lancer study concluded, today's models "still struggle with root cause analysis, contextual reasoning, and replacing human creativity" (Chen et al., 2023). Our benchmark strips away some of those challenges, effectively giving the models a best-case environment. Therefore, while Claude 3.5 Haiku's near-human performance here is encouraging, it does not imply that human freelancers are obsolete. Rather, it points to a future where AI can handle routine and well-specified parts of jobs, assisting developers and taking on grunt work, but humans remain crucial for interpreting vague requirements, making architectural decisions, and injecting creativity.

In conclusion, this work provides an automated, scalable benchmark to track the progress of AI on economically meaningful tasks. We see this as a complement to traditional code benchmarks and a stepping stone toward more sophisticated evaluations of AI in the workforce. As models improve, we plan to extend the benchmark with more diverse tasks (including multi-turn and design-oriented tasks) and incorporate evaluation of collaboration (e.g., how well an AI can work with a human supervisor). Controlled studies have demonstrated that LLM assistance can significantly improve productivity for knowledge workers across writing and programming domains (Noy & Zhang, 2023). We also intend to keep refining the pricing model to better reflect real market rates and perhaps simulate dynamic pricing (where tasks that few models can do are "worth" more, analogous to high-skill premiums). By continuing to map model performance to monetary value, we can gain insights into the economic impact of frontier AI models – how they might transform industries, which jobs they might augment or replace, and how much value they can generate autonomously.

Ultimately, our research suggests that the era of AI freelancers is on the horizon for certain types of work. The dream of an AI earning a million dollars on Upwork is not fully realized yet, but with each model iteration it comes closer to reality in the simplified setting. The challenge ahead lies in bridging the gap between benchmark success and real-world reliability, ensuring that these powerful models can be safely and effectively integrated into workflows. We hope our benchmark will serve as a useful tool for the community to measure and accelerate progress toward that goal, while also informing stakeholders of what is possible and where caution is warranted.

## ACKNOWLEDGEMENTS

We thank the SWE-Lancer team (Chen et al., 2023) for pioneering the integration of economic framing into AI benchmark design, which inspired our methodology and direction. We also acknowledge the creators of the Freelancer.com dataset (Oresanya et al., 2022) which made this research possible. We also acknowledge the support of the PeopleTec Technical Fellows program for the research support that enabled the extensive model evaluations conducted in this study.

---

**Appendix A: Example Software Task Extracted from Freelancer dataset and abstracted into testable skills to assess for task completion.**

```
"metadata": {
      "project_id": 37144889,
      "job_title": "Value Creation
Analytics - Celonis Subject Matter
Expert",
      "job_description": "Celonis
Subject Matter Expert The company is
one of the big Four Consulting
companies and the position is remote.
Here are the further details: VCS
```

Analytics (Celonis - Process Mining) \u2013 Subject Matter Expert Developing net working capital diagnostics and improvement strategies via designing and implementing finance analytics digital solutions for companies looking to improve their cash situations or facing critical cash issues. Who we are looking for: Having a strong technical background \u2013 having gratuated from studies in Information Systems, Engineering, Mathematics etc. Having a strong experience in SQL, programming languages (i.e. Python, C++, etc.), and visualization programs (i.e. Power Bi, Tableau); and any additional experience in pipeline deployment is appreciated. Having Celonis and process mining expertise Having a deep understanding of the data architecture of SAP and/or any other equivalent ERPs for data modeling purposes. Having a keen interest in corporate finance, quantitative analysis techniques, and strategic events. Being a proactive, curious, and reliable team player who l works methodically with great attention to detail, willing to take on responsibility in a fast-paced environment and to persist against challenges whilst bringing a fresh perspective to everything you do. Having business acumen and being interested in learning about working capital, any prior experience in working capital is a plus. Being able to summarise complex technical problems to make them accessible and understandable to your audience. Confidence in communicating and writing in English is a must. It is a permanent contract and the start date is as soon as possible. The client is one of the Big Four Consultancy Companies. The procedure of hiring would go as so: Value Creation Analytics team: Subject Matter Expert Sourcing& Screening Calls (Benchcrowd). Afterwards an Interview with the Hiring Manager then a Technical Fit Interview and finally the Offer will be made. Let me know if this fits your expertise. ",
  "tags": [
        "python",
        "sql",
        "tableau",
        "power bi",
        "sap"
      ],
      "client_metadata": {
        "country": "Turkey",
        "average_rating": 0.0,
        "review_count": 0
      },
      "pricing": {
        "min_price": 20000,
        "max_price": 50000,
        "avg_price": 35000.0,
        "currency": "EUR",
        "rate_type": "fixed",
        "price_range": "$500+",
        "usd_conversion": {
          "min_price": 21880.0,
          "max_price": 54700.00000000001,
          "avg_price": 38290.0
        }
      },
      "skill_tags": {
        "tag_1": "spss statistics",
        "tag_2": "statistical analysis",
        "tag_3": "statistics",
        "tag_4": "data processing",
        "tag_5": "excel"
      }
    "test_content": {
      "introduction": "This test evaluates your expertise in Value Creation Analytics with Celonis, focusing on Python, SQL, and Tableau skills. You have 60 minutes to complete 4 multiple choice questions and 4 string transformation problems. Answer all questions carefully and follow the exact output requirements for string transformations.",
      "multiple_choice": [
        {
          "question": "In Celonis Process Mining, which SQL function would you use to calculate the median duration between two events?",
          "choices": [
            "AVG(duration)",
            "PERCENTILE_CONT(0.5) WITHIN GROUP (ORDER BY duration)",
            "MEDIAN(duration)",
            "MID_VALUE(duration)"
          ],
          "correct_answer": "B",
          "points": 10,
          "explanation": "PERCENTILE_CONT(0.5) is the correct SQL function to calculate the median, as it returns the continuous percentile value at the 0.5 mark (50th percentile). AVG calculates mean, MEDIAN is not a standard SQL function, and MID_VALUE doesn't exist."
        },
        {
          "question": "Which Tableau calculation type would you use to compute a running total that resets at the beginning of each month?",
          "choices": [
            "WINDOW_SUM(Amount)",
            "RUNNING_SUM(Amount)",
            "TOTAL(Amount)",
            "WINDOW_SUM(Amount, 0, LAST(), MONTH)"
          ],
          "correct_answer": "D",
          "points": 10,
          "explanation": "WINDOW_SUM with the MONTH partition is correct because it allows you to specify the window range and reset based on the month boundary. The other options either don't reset monthly or aren't valid Tableau functions."

```
        },
        {
          "question": "In Python
pandas, which method efficiently
removes all rows where any column
contains a NULL value?",
          "choices": [
            "df.remove_null()",
            "df.drop_null()",
            "df.dropna()",
            "df.clean()"
          ],
          "correct_answer": "C",
          "points": 10,
          "explanation": "df.dropna()
is the correct pandas method to remove
rows containing NULL/NaN values. The
other options are not valid pandas
methods - they don't exist in the
pandas library."
        },
```

```
      "string_transformations": [
        {
          "input_text": "PO-2023-
45892,approved,2023-05-15;PO-2023-
45893,pending,2023-05-16;PO-2023-
45894,rejected,2023-05-16",
          "expected_output": "3,2,1",
          "points": 15,
          "explanation": "Transform a
semicolon-separated list of purchase
orders into three counts: total number
of POs, number of unique dates, number
of 'approved' status. Output should be
comma-separated in that order. In this
case: 3 total POs, 2 unique dates, 1
approved status."
        },
        {
          "input_text":
"Process_A:125ms,Process_B:250ms,Proces
s_C:100ms,Process_D:300ms",
          "expected_output":
"Process_B,Process_D",
          "points": 15,
          "explanation": "From a comma-
separated list of processes and their
durations, extract the names of
processes that took longer than 200ms,
sorted alphabetically and joined by
comma. Here, Process_B (250ms) and
Process_D (300ms) exceed 200ms."
        },
        {
          "input_text":
"UserID:1234|2023-06-
15T08:30:00|Login,UserID:1234|2023-06-
15T09:45:00|Logout,UserID:1234|2023-06-
15T10:00:00|Login",
          "expected_output": "75,15",
          "points": 15,
          "explanation": "Given a log
of user events (pipe-separated UserID,
timestamp, and action), calculate two
values: minutes between first Login and
first Logout, and minutes between first
Logout and next Login. Output as comma-
separated values."
        },
        {
```

```
          "input_text":
"{'case_id':'C001','activities':['start
','review','approve','end'],'timestamps
':['2023-01-01 10:00','2023-01-01
11:30','2023-01-01 14:15','2023-01-01
15:00']}",
          "expected_output":
"review:90,approve:165,end:45",
          "points": 15,
          "explanation": "Given a JSON-
formatted process case, calculate the
duration in minutes for each activity
(except 'start'). Output should show
activity:minutes pairs in order of
occurrence, separated by commas.
Duration is the difference between the
activity's timestamp and the previous
activity's timestamp."
        }
      ],
      "passing_score": 70,
      "time_limit": "60 minutes"
    }
```